# A New Optimization Approach Based on Rotational Mutation and Crossover Operator

Masoumeh Vali

Department of Mathematics, Dolatabad Branch, Islamic Azad University, Isfahan, Iran
E-mail: vali.masoumeh@gmail.com

**Abstract**

Evaluating a global optimal point in many global optimization problems in large space is required to more calculations. In this paper, there is presented a new approach for the continuous functions optimization with rotational mutation and crossover operator.
This proposed method (RMC) starts from the point which has best fitness value by elitism mechanism and after that rotational mutation and crossover operator are used to reach optimal point. RMC method is implemented by GA (Briefly RMCGA) and is compared with other well-known algorithms such as: DE, PGA, Grefensstette and Eshelman[15,16] and numerical and simulating results show that RMCGA achieve global optimal point with more decision by smaller generations.

1. Introduction

This paper concerned with the simple-bounded continuous optimization problem as follows:
f $(x_1, x_2, \ldots, x_m)$ where each $x_i$ is a real parameter so that $a_i \leq x_i \leq b_i$ for some constants $a_i$ and $b_i$ and $i = 1, 2, \ldots, m$.

This problem has widespread applications including optimization simulating models, fitting nonlinear curve to data, solving system of nonlinear, engineering design and control problem, and setting weights on neural networks.

Since GA provides a comprehensive search methodology for optimization, this problem is implemented by GA for more optimal performance. In global optimization scenarios, GAs often manifests their strengths: efficiency, parallelizable search; the ability to evolve solutions with different dimension; and a characterized and controllable process of innovation.

In this paper is presented a new approach for simple-bounded continuous optimization problem that is called RMC. In other words, RMC is a new approach for finding global optimal point (max/min) by simple algebra, without derivation and based on search.

This method, at first, search the point has the best fitness in its search space and after that by using rotational mutation and crossover operator finds the global optimal point.

This paper starts with the description of related work in section 2. Section 3 gives the model and problem definition of RMC. In section 4, text problems of RMC method is implemented. In section 5, Schemata Analysis for RMCGA is present. Evaluation by De Jong's functions and the compression of RMCGA with the other methods (DE, PGA, Grefensstette and Eshelman) for De Jong Functions are shown in sections 6. The discussion ends with a conclusion and future trend.

## 2. Related Work

In the early 1960s and 1970s, new search algorithms were initially proposed by Holland, his colleagues and his students at the University of Michigan. These search algorithms which are based on nature and mimic the mechanism of natural selection were known as Genetic Algorithms (GAs) [1, 2, 3, 4, 5, 6, 7]. Holland in his book "Adaptation in Natural and Artificial Systems" [1] initiated this area of study. Theoretical foundations besides exploring applications were also presented.
As a matter of fact, "Genetic algorithms' functionality is based upon Darwin's theory of evolution through natural and sexual selection." [6], they mimic biological organisms [3]. In GAs a solution to the problem is represented as a genome (or chromosome) [1, 3, 4].

Pratibha Bajpal and Manojkumar [9] have proposed an approach to solve Global Optimization Problems by GA and obtained that GA is applicable to both continuous and discrete optimization problems.

Hayes and Gedeon [10] considered infinite population model for GA where the generation of the algorithm corresponds to a generation of a map. They showed that for a typical mixing operator all the fixed points are hyperbolic.

Gedeon et al. [11] showed that for an arbitrary selection mechanism and a typical mixing operator, their composition has finitely many fixed points.

Qian et al.[12] proposed a GA to treat with such constrained integer programming problem for the sake of efficiency. Then, the fixed-point evolved (E)-UTRA PRACH detector was presented, which further underlines the feasibility and convenience of applying this methodology to practice.

Devis Karaboga and Selcuk. [15] is proposed new heuristic approach with deferential evaluation (DE) for finding a true global minimum regardless of the initial parameter values, fast convergence using similar operators crossover, mutation and selection.

## 3. Model and Problem Definition of RMC

In this section, at first, model of RMC is expressed.

Suppose f($x_1, x_2, x_3 \ldots, x_n$) with constraints $a_i \leq x_i \leq b_i$ for i=1, 2,…, n. The serial algorithm RMC is as follows:

**Step 1:** Draw the diagrams for $x_i = b_i$ and $x_i = a_i$ for i=1, 2,…, n.

**Step 2:** Consider the vertex of this polytope which has best fitness among other vertexes and call it S.

**Step 3:** Put $S' = f(S)$, $\vec{r_0}$=(1,1,…,1), (1,-1,1,…,1),…, or (-1,-1,…,-1) and $r_n = \alpha \vec{r_0}, = 0.1n$, n=1,2,…,10.

**Step 4:** Move the point 'S' with length of α in direction of $\vec{r_0}$ vector (notice: the direction of $\vec{r_0}$ vector must be inside the search space, also the α measurement depends on problem precision). The endpoint of vector $r_n$ is called P.

**Step 5:** If f (P) better than $S'$, then put S=P and go to step 8 else go to step 6.

**Step 6:** Put $\beta = 0.1, 0.25, \ldots, e_n = (1,1,\ldots,1), (1,-1,1,\ldots,1), \ldots,$ and $(-1,-1,\ldots,-1), P = \beta e_n$.

**Step 7:** If f (P) is better than $S'$, then S=P else go to step 8.

**Step 8:** If P is in search space, go to step 9 else go to stop.

**Step 9:** Halve the adjacent sides of polytope. The result is the production of new points.

**Step 10:** Select point has best fitness among the new points produced from step 9 and point 'S'. Then, put S the point which has the best value and S' fitness of point S then go to step 3 and repeat this trend.

## 4. Implementing Test problems by RMC

### 4.1. Test problem1:

The minimization problem is formulated as follows:

$$\min_{x \in \mathbb{R}^2} f(x) = x_1^2 + x_2^2 - 18 \cos x_1 - 18 \cos x_2 \qquad (5$$

$$-1 \leq x_i \leq 1, \quad i = 1,2$$

Its global minimum is equal to -2 and the minimum point is at (0,0).
The process of achieve to optimal global point by RMC is shown following figure.

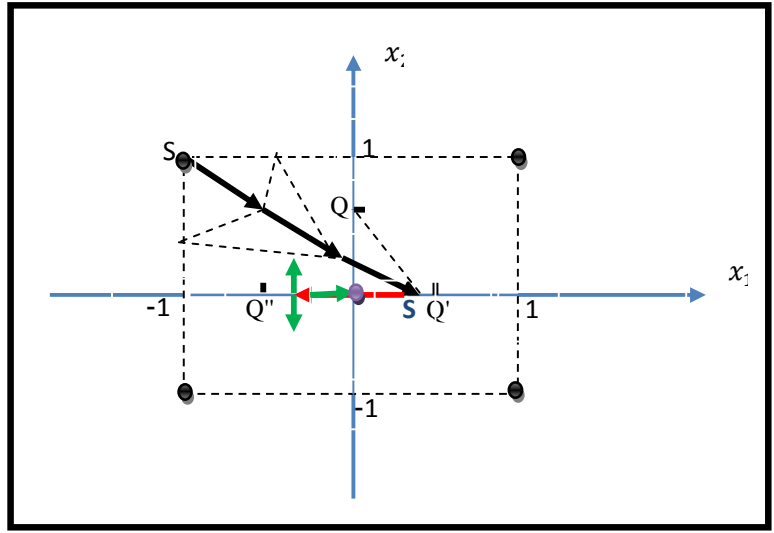

Figure 1: The Performance RMC

Note that f(1,1) = f(-1,1) = f(1,-1) = f(-1,-1) so it's not important that from which side start. As you can see, the first two mutations improve the fitness of problem- The fitness value of mutated points are better than the fitness value of points produced by crossover - but the fitness of the third mutation cannot improve the fitness of problem so use crossover operator and produce Q and Q' points.Then among S, Q and Q' select S which has best fitness and after that rotational mutation and crossover operator is used for achieve global optimal point.

### 4.2. Test problem2:

The Goldstein-Price function [GP71] is a global optimization test function.
function definition:
$$f_{\text{Gold}}(x_1, x_2) = \left(1 + (x_1 + x_2 + 1)^2 . (19 - 14x_1 + 3x_1^2 - 14x_2 + 6x_1 x_2 + 3x_2^2)\right).$$
$$\left(30 + (2x_1 - 3x_2)^2 . (18 - 32x_1 + 12x_1^2 + 48x_2 - 36x_1 x_2 + 27x_2^2)\right)$$
$$-2 \leq x_i \leq 2, i = 1,2$$

Global minimum:
$$f(x_1, x_2) = 3; \ (x_1, x_2) = (0, -1)$$

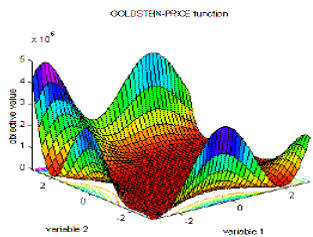

Figure 2: Figure of the Goldstein-Price

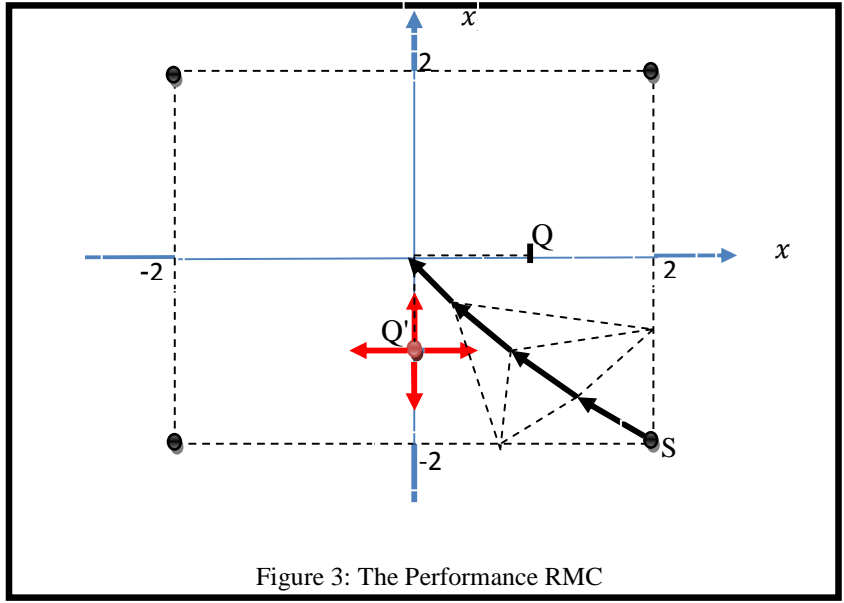

Figure 3: The Performance RMC

As regards the fitness value of points (2,-2) and (-2,-2) is equal and better than the fitness value of points (2, 2) and (-2, 2) so we start from point (2,-2). As you can see, the first third mutations improve the fitness of problem- The fitness value of mutated points are better than the fitness value of points produced by crossover - but the fitness of the forth mutation cannot improve the fitness of problem so use crossover operator and produce Q and Q' points and select Q' which has the best fitness. After that use rotational mutation to achieve point which the improvement is better but this point didn't find so point Q' is chosen as the global point optimal.

5. Schema Analysis RMCGA

This schema (Figure 4) starts from the vertex (offspring) which has best fitness value that is called 'S'. Then it sets initial point of $\vec{r_0}$ vector at 'S' and mutates offspring 'S' in direction of $\vec{r_0}$ vector with length of mutation α. ( notice that the direction of $\vec{r_0}$ vector must be inside the search space, also the α measurment depends on problem precision.) . This mutated offspring is calledP.

If fitness value of offspring P was better than the fitness value of offspring S we would use crossover operator for adjacent edges of offspringP. Otherwise by using rotational mutation by $\vec{e_n}$ vector, we would search an offspring-with better fitness value in comparison with'S'. Then we make a crossover. After that we select an offspring with the better fitness value –between the mutated offspring and offspring's which were generated crossover operator. We mutate and make a crossover on it again. We repeat this action while the mutated offspring doesn't get out of search space. Eventually, we would select the last new produced offspring- inside the search space- as the global optimization point which would be fixed point of our question.

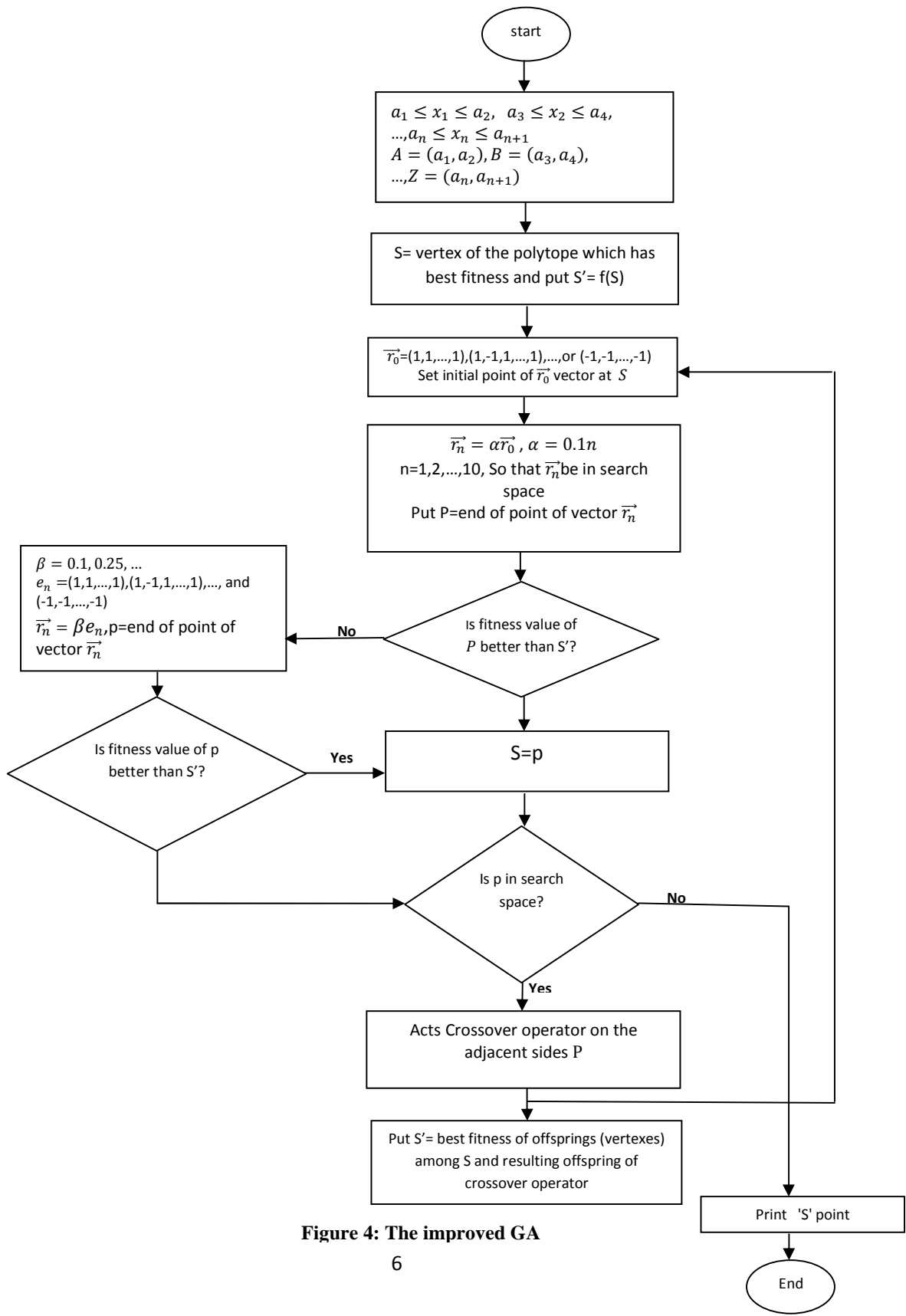

**Figure 4: The improved GA**

## 6. Evaluation

In this section, Definition of De Jong's Functions, Numerical results of RMCGA on De Jongs' functions and finally simulating results and the compression of RMCGA with the other methods such as DE, PGA, Grefensstette and Eshelman[15, 16] are shown.

### 6.1 De Jong's Functions

In this section, Definition of De Jong's Functions (F1 to F5) and initial population RMCGA which depends on the dimensional space are shown in Table 1.

Table 1: De Jong's Functions

| Function Number | Function | Limits | Dim. | Initial Population |
|---|---|---|---|---|
| F1 | $\sum_{i=1}^{3} x_i^2$ | $-5.12 \leq x_i \leq 5.12$ | 3 | 8 |
| F2 | $100.(x_1^2 - x_2)^2 + (1 - x_1)^2$ | $-2.048 \leq x_i \leq 2.048$ | 2 | 4 |
| F3 | $30. + \sum_{i=1}^{5} \lfloor x_j \rfloor$ | $-5.12 \leq x_i \leq 5.12$ | 5 | 32 |
| F4 | $\sum_{i=1}^{30} (ix_i^4 . + Gauss\,(0,1))$ | $-1.28 \leq x_i \leq 1.28$ | 30 | |
| F5 | $\dfrac{1}{0.002 + \sum_{i=0}^{24} \frac{1}{i + \sum_{j=0}^{1}(x_j - a_{ij})^6}}$ | $-65.536 \leq x_i \leq 65.536$ | 2 | 4 |

### 6.2. Numerical Results

In this section, the experimental results of RMCGA on the five problems of De Jong (F1 to F5) [De Jong, 1975] in Table 2 are shown. Furthermore, results were saved for the best performance (BP) which BP is the best fitness of the objective function obtained over all function evaluations. At last, standard deviation (SD) is calculated and measured with final answer of De Jong function. The following parameters were evaluated in the following table.
rotational mutation size (RMS), number of iterations rotational mutation (TRM), number of iterations crossover operator (TC).

Table 2: The average number of generations.

| Algorithm | De Jong's Function | RMS | TRM | TC | Best Point | BP | SD |
|---|---|---|---|---|---|---|---|
| **RMCGA** | F1 | 0.1 | 55 | 102 | (0,0,0) | 0 | 0 |
| **RMCGA** | F2 | 0.1 | 15 | 20 | (1,1) | 0 | 0 |
| **RMCGA** | F3 | 0.1 | 4 | 1 | (-5.12,-5.12,-5.12,-5.12,-5.12) | 0 | 0 |
| **RMCGA** | F4 | 0.1 | 15 | 21 | $(\underbrace{0,0,\ldots,0}_{30})$ | Depend on $\eta$ | 0 |
| **RMCGA** | F5 | 0.1 | 340 | 670 | (-32,-32) | 0 | 0 |

## 6.3. Simulating Results

Dervis Karaboga and Selcuk Ökdem[15] introduced new method DE for finding global optimization problems. They compared DE method with other methods: PGA, Grefensstette and Eshelman [16] and showed that DE algorithm works much better than other methods[16].
In order to obtain the average results, PGA, Grefensstette and Eshelman algorithms were run 50 times; the DE algorithm was run 1000 times and RMCGA 200 times for each function.

Table 3: The average number of generations

| **Algorithms** | **F1** | **F2** | **F3** | **F4** | **F5** |
|---|---|---|---|---|---|
| **PGA($\lambda = 4$)** | 1170 | 1235 | 3481 | 3194 | 1256 |
| **PGA($\lambda = 8$)** | 1526 | 1671 | 3634 | 5243 | 2076 |
| **Grefensstette** | 2210 | 14229 | 2259 | 3070 | 4334 |
| **Eshelman** | 1538 | 9477 | 1740 | 4137 | 3004 |
| **DE(F: RandomValues)** | 260 | 670 | 125 | 2300 | 1200 |
| **RMCGA** | 157 | 35 | 5 | 36 | 1010 |
| **PNG** | 1.656 | 19.142 | 25 | 63.888 | 1.188 |

As you can see in Table 3, the convergence speed RMCGA achieves the optimal point with more decision by smaller generation. Furthermore, the most significant improvement is with F4 since the proportion of the number of generations (PNG) about $\frac{2300}{36} \cong 64$ times smaller than the average of the number of generations DE algorithm.

### 7. Conclusion

RMCGA is a new method for finding the true optimal global optimization is based on rotational mutation and crossover operator. In this work, the performance of the RMCGA has been compared to that of some other well known GAs. From the simulation studies, it was observed that RMCGA achieve the optimal point with more decision by smaller generation. Therefore, RMCGA seems to be a promising approach for engineering optimization problems.